\documentclass[a4paper]{article}

\usepackage{INTERSPEECH2021}
\usepackage{subfig}
\usepackage{bm}
\usepackage{float}
\usepackage{amsfonts}
\usepackage{diagbox}
\usepackage{color}
\usepackage{url}
\usepackage{cite}
\usepackage{makecell}

\def\M{\mathrm{M}}
\def\FF{\mathrm{FF}}

\def\bQ{\bm{Q}}
\def\bK{\bm{K}}
\def\bV{\bm{V}}

\def\A{\mathrm{A}}
\def\H{\mathrm{H}}
\def\Q{\mathrm{Q}}
\def\K{\mathrm{K}}
\def\V{\mathrm{V}}

\title{Stochastic Attention Head Removal: A simple and effective method for improving Transformer Based ASR Models}

\name{Shucong Zhang, Erfan Loweimi, Peter Bell, Steve Renals\thanks{SZ was supported by a PhD Studentship funded by Toshiba Research Europe Limited.}}
\address{Centre for Speech Technology Research, University of Edinburgh, Edinburgh, UK}
\email{s1603602@sms.ed.ac.uk, \{e.loweimi, peter.bell, s.renals\}@ed.ac.uk}

\begin{document}

\maketitle
\begin{abstract}

Recently, Transformer based models have shown competitive automatic speech recognition (ASR) performance. One key factor in the success of these models is the multi-head attention mechanism. However, for trained models, we have previously observed that many attention matrices are close to diagonal, indicating the redundancy of the corresponding attention heads.  We have also found that some architectures with reduced numbers of attention heads have better performance. Since the search for the best structure is time prohibitive, we propose to randomly remove attention heads during training and keep all attention heads at test time, thus the final model is an ensemble of models with different architectures. The proposed method also forces each head independently learn the most useful patterns. We apply the proposed method to train Transformer based and Convolution-augmented Transformer (Conformer) based ASR models. Our method gives consistent performance gains over strong baselines on the Wall Street Journal, AISHELL, Switchboard and AMI datasets. To the best of our knowledge, we have achieved state-of-the-art end-to-end Transformer based model performance on Switchboard and AMI.

\end{abstract}

\section{Introduction}
\label{sec:intro}
In self-attention based models such as Transformers \cite{vaswani2017attention}, each self-attention layer uses multi-head attention to capture a set of different inputs representations, and have given competitive ASR results \cite{Dong2018Speech-transformer, pham2019veryDeep, karita2019comparative, han2019multi,jiang2020further, tian2020synchronous, bai2020laso, luo2020simplified, gulati2020conformer}. However, we previously observed that not all of the attention heads are useful \cite{zhang2021usefulness}. We found that in trained Transformer ASR models, the attention matrices of some attention heads are close to the identity matrices, indicating that the attention mechanism is just an identity mapping. Thus, the input sequence (after a linear transformation) can be directly used as the output of the corresponding attention head.


This observation leads to a question: if there are redundant attention heads in trained Transformer models, then when we train a model from scratch, can we remove some attention heads without harming the model performance? In our experiments, we found some architectures with reduced numbers of attention heads offer better accuracies. However, searching for the best structure is time consuming. To address this, we propose to randomly remove attention heads during training, and keep all the attention heads at test time. Through this method, the trained model is an ensemble of different architectures. 

Our method also enables the networks to better utilize the representation power of the multi-head attention architecture. The redundancy of attention heads in the baselines indicates some heads learn the most crucial patterns while other heads merely extract inessential features which can be disregarded (this is also observed in \cite{voita2019analyzing}). In contrast, our method forces all attention heads to learn crucial features -- since heads are randomly removed during training, the model cannot only rely on some specific heads for extracting key features. Thus, the proposed method forces each head to learn useful information and enables the model to better use the representation power of the multi-head attention structure. Our experimental results and analysis on attention matrices support our arguments. 

We employ the proposed method to train Transformer and Conformer \cite{gulati2020conformer} ASR models. Our method offers consistent accuracy improvements on datasets of different ASR scenarios, including Wall Street Journal (WSJ) \cite{paul1992wsj}, AISHELL \cite{bu2017aishell}, Switchboard (SWBD) \cite{godfrey1992switchboard} and AMI \cite{renals2007recognition}. On SWBD and AMI, to the best of our knowledge, we have achieved state-of-the-art (SOTA) performance for end-to-end Transformer based models. 


\section{Related Work}

At first sight, the proposed method appears to be similar to dropout \cite{srivastava2014dropout}; however the objectives of these two methods are different. Dropout randomly drops units in neural networks to prevent the co-adaptation between these units. In this work, we randomly remove entire attention heads, motivated by the observation that some architectures with reduced numbers of attention heads can lead to better performance, but avoiding the need for a time-consuming search for the best architecture.  Instead, we effectively train an ensemble of different structures.  Furthermore, we observe our method forces the attention heads which are redundant in the baseline systems to learn useful patterns. We also have employed dropout in our experiments and found it is complementary to our method.

Previous works have proposed to use a development set \cite{michel2019are} or a trainable hard gate head \cite{voita2019analyzing} to drop attention heads. These methods remove attention heads in a static manner -- once an attention head is removed during training, it will never be present in the structure. Also, little performance gain is obtained by these methods. Our proposed method removes attention heads randomly and consistently yields better accuracies. 

To alleviate the gradient vanishing problem and build very deep networks, previous works have proposed to drop self-attention layers randomly or following a schedule \cite{pham2019veryDeep,fan2019reducing}. Rather than dropping self-attention layers, our method removes individual attention heads randomly. The motivation of our method also differs from the previous works. Thus, the proposed method is complementary with these previous works.




Peng et al~\cite{peng2020mixture} proposed to train subsets that contain $h-1$ elements of the $h$ attention heads. During testing, the output is a linear combination of the outputs of these subsets. In our work, the attention heads are randomly removed. Therefore, the number of trainable attention heads changes dynamically.

Zhou et al~\cite{zhou2020scheduled} proposed a similar method of training self-attention models for natural language processing (NLP). However, the related work does not analyse where the benefits of such methods come from, whilst in contrast, we independently derive our method based on our previous work which studies the usefulness of the attention heads \cite{zhang2021usefulness}. Additionally, in this work, our detailed analysis of the proposed method reveals why dynamically removing attention heads  is helpful. Furthermore, the previous work relates to NLP while our method focus on models which are specifically designed for ASR.


\section{Transformer Based Models}
In this section, we briefly review Transformer \cite{vaswani2017attention} and Conformer \cite{gulati2020conformer}. The basic building block for Transformer based models is self-attention \cite{vaswani2017attention}. A self-attention layer takes a 3-tuple of sequences as its input. Each self-attention layer has a multi-head attention (MHA) component and a feed-forward component. Each attention head $\A_{i}$ in the MHA uses a linear transformation to map the input sequences to a Query/Key/Value triplet, which is denoted as $(\bQ,\bK,\bV)$. Then, it uses the dot product to compute the similarity between each element of the $\bK$ sequence and each element of the $\bQ$ sequence. Using these similarities as weights, each element of the output sequence of the attention head is the weighted sum of the $\bV$ sequence. Thus, an attention head $\mathrm{A}_{i}$ can be described as:
%
%
\begin{align}
\mathrm{A}_{i}(\bm{X}^\Q, \bm{X}^\K, \bm{X}^\V) &= \mathrm{softmax}(\frac{\bm{Q}_{i} \bm{K}_{i}^T }{\sqrt{d^{\mathrm{K}}}})\bm{V}_{i}\\
(\bm{Q}_{i} ,\bm{K}_{i} ,\bm{V}_{i}) &= (\bm{X}^\Q \bm{W}_{i}^\Q, \bm{X}^{\K}\bm{W}_{i}^{\K},\bm{X}^\V \bm{W}_{i}^\V)
\end{align}
where $ \bm{X}^\Q \in \mathbb{R}^{ n \times d^{\M}}$, $\bm{X}^\K, \bm{X}^\V \in \mathbb{R}^{ m \times d^{\M}}$ are inputs and $m,n$ denote the lengths of the input sequences; $\bm{W}_{i}^\Q, \bm{W}_{i}^\K \in \mathbb{R}^{ d^{\M} \times d^{\K}} $ and  $\bm{W}_{i}^\V \in \mathbb{R}^{ d^{\M} \times d^{\V}}$ are trainable matrices. 

The multi-head attention $\mathrm{MHA}$ is a linear combination of the outputs of the $h$ attention heads $(\A_1,\A_{2}, \cdots, \A_h)$, which can be described as:
%
%
\begin{equation}
    \mathrm{MHA}(\bm{X}^\Q, \bm{X}^\K, \bm{X}^\V) = (\A_{1},\A_{2},\cdots,\A_{h}) \ \bm{U}^\H,
\end{equation}
where $\bm{U}^\H$ is a trainable matrix and $\bm{U}^\H \in \mathbb{R}^{d^{\H} \times d^{\M}}, d^{\H} =h \times d^{\V}$. A residual connection links the the output of the $\mathrm{MHA}$ and the query sequence $\bm{X}^\Q$:
\begin{equation}
    \bm{X}' = \bm{X}^\Q + \mathrm{MHA}(\bm{X}^\Q,\bm{X}^\K,\bm{X}^\V)
\end{equation}
Finally, there is a feed-forward component upon the MHA: 
\begin{equation}
    \bm{Y} = \bm{X}' + \mathrm{ReLU}(\bm{X}'\bm{S} +\bm{b})\bm{Z}+\bm{r}
\end{equation} where $\bm{S} \in \mathbb{R}^{d^{\M}\times d^{\FF}}$, $\bm{Z} \in \mathbb{R}^{d^{\FF} \times d^{\M}}$, $\bm{b} \in \mathbb{R}^{d^{\FF}}$ and $\bm{r} \in \mathbb{R}^{d^{\M}}$ are trainable matrices and vectors.

Transformer is a self-attention based model, which has a self-attention encoder to encode the input sequences and a self-attention decoder to decode the encoded sequences. For each self-attention layer in the encoder,  $\bm{X}_e^\Q = \bm{X}_e^\K=  \bm{X}_e^{\V}$ and each sequence is the speech signal or its hidden representation from the previous layer. Each self-attention layer in the decoder has two MHA components and one feed-forward component. The first MHA attends the previous output tokens or the outputs of the previous layer. Then, the second MHA looks at the encoded speech signal, $\bm{X}_s^\K=  \bm{X}_s^{\V}$ are the final hidden representations of the input speech sequence, and $\bm{X}_s^\Q$ is the output of the previous MHA (with the residual connection). 

Since self-attention encodes sequences through attention mechanisms, it can model dependencies within any range. However, previous works have shown it is less effective in capturing local information \cite{zhang2021usefulness, wu2020lite}. Nevertheless, acoustic events usually happen within short periods and thus local information is essential for ASR tasks. Since convolution neural networks (CNNs) are suitable for capturing local context,  Gulati et al \cite{gulati2020conformer} propose Conformer, which augments self-attention layer encoders with CNNs. A Conformer layer can be described as:

\vspace{-5mm}

\begin{align}
   & \bm{\tilde{X}} = \bm{X} + \frac{1}{2} \FF(\bm{X}) \\ 
   & \bm{X'} = \bm{\tilde{X}} +  \mathrm{MHA}(\bm{\tilde{X}},\bm{\tilde{X}},\bm{\tilde{X}}) \\
   & \bm{X''} = \bm{X'} + \mathrm{Conv}(\bm{X'}) \\
   & \bm{Y} = \mathrm{LayerNorm}(\bm{X''} + \frac{1}{2} \FF(\bm{X''})) 
\end{align}
where $\FF$ denotes the feed-forward component \cite{gulati2020conformer}, $\mathrm{Conv}$ denotes the convolution component \cite{gulati2020conformer} and $\mathrm{LayerNorm}$ denotes layer normalisation \cite{ba2016layer}. The half-step feed-forward networks $\frac{1}{2} \FF$ have resulted in better accuracies compared to the vanilla feed-forward network \cite{gulati2020conformer}. In the original Conformer model, long short-term memory (LSTM) \cite{hochreiter1997long} is used as the decoder. In this paper, we use ``Conformer'' to denote the \textbf{Conformer Encoder--Transformer Decoder} structure. 


\section{Stochastic Attention Head Removal}

\begin{figure}[t]
\centering
\subfloat[Baseline]{\includegraphics[width=0.42\columnwidth]{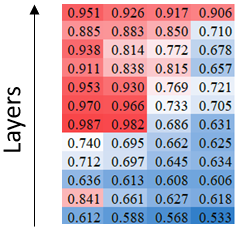}}
\qquad
\subfloat[remove  topmost]{\includegraphics[width=0.42\columnwidth]{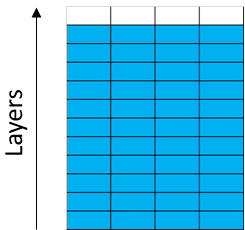}} 
\qquad
\subfloat[remove $\mathrm{diag}>0.95$]{\includegraphics[width=0.42\columnwidth]{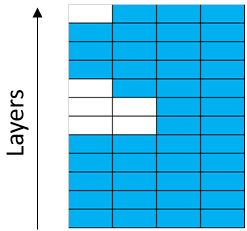}}
\qquad
\subfloat[remove $\mathrm{diag}>0.90$]{\includegraphics[width=0.42\columnwidth]{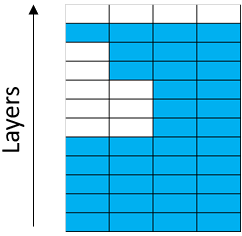}}
\caption{The tested encoder architectures. The baseline has 12 self-attention layers. Each non-white cell represents an attention head. Each white cell denotes there is no attention head. When all the attention heads of a layer are removed (four white cells), the feed-forward component of that layer is preserved. (a) heatmap of the averaged diagonality of each attention head. (b-d) structures where some attention heads are removed. } 
\label{fig:static}
\end{figure}

We introduce the stochastic attention head removal (SAHR) strategy in this section. In our previous work \cite{zhang2021usefulness}, we found in trained Transformer ASR models, there are heads which only generates nearly diagonal attention matrices. The high diagonality indicates these heads are merely identity mappings and thus they are redundant. We further investigate if removing the heads with high diagonality and retrain the model from scratch will impact the performance. We use WSJ as the dataset and the experimental setups are presented in Section~\ref{sec:exp}. For the baseline model, following our previous work, we plot the heatmap of the averaged diagonality \cite{zhang2021usefulness} on WSJ eval92 for each head in every encoder layer. Then, we remove the heads whose averaged diagonality are above a threshold and retrain the model. We also test removing all the heads in the uppermost encoder layer (near output), since all the heads in this layer have high diagonality. Figure~\ref{fig:static} shows the heatmap and the structures with the corresponding head removal strategy.

Table~\ref{tab:stat} shows the character error rates (CERs) for each structure. We observe some architectures outperform other structures. However, the search for the best architecture is time-consuming. To address this, we propose to randomly remove attention heads during training and keep all heads at test time. The proposed method can be viewed as training different architectures and using the ensemble of these trained architectures as the final model at test time. When processing a training example, each attention head has a probability $q$ of being removed, where $q$ is set as a hyper-parameter. If a head is kept, then the output of that head is scaled by $\frac{1}{1-q}$: 

\vspace{-5mm}

\begin{equation}
    \mathrm{A}_{i}(\bm{X}^\Q, \bm{X}^\K, \bm{X}^\V) = \frac{1}{1-q} \times
    \mathrm{softmax}(\frac{\bm{Q}_{i} \bm{K}_{i}^T }{\sqrt{d^{\mathrm{K}}}})\bm{V}_{i}
\end{equation}
At test time, the scale factor $\frac{1}{1-q}$ is removed and all the attention heads are always present. Thus, the expected output of the attention head is the same during training and testing. During training, if all the attention heads of a layer are removed for an utterance, then that layer becomes a feed-forward layer. The stochastic attention head removal strategy is applied for both the encoder and the decoder. Table~\ref{tab:stat} shows the preliminary results of the our method. In Section~\ref{sec:exp}, we further test applying the proposed method to train both Transformer and Conformer ASR models on other larger and more challenging datasets.

\begin{table}[t]
\caption{CERs on WSJ for different removal strategies. The structures of the static removal methods are shown in Figure~\ref{fig:static}. \#heads/ denotes the number of attention heads.}
\vspace{-2.5mm}
\label{tab:stat}
\centering
\begin{tabular}{l|c|c}
\hline
Removal Method    & \# Heads  & eval 92/dev 93       \\ \hline \hline
None (baseline) & 48               & 3.5/4.6          \\
Remove topmost  heads      & 44    & \textbf{3.4}/\textbf{4.5}          \\
Remove heads $\mathrm{diag}>0.95$        & 42              & \textbf{3.4}/4.6          \\
Remove heads $\mathrm{diag}>0.90$       & 36              & 3.5/4.7 \\ \hline 
\\[-3.0mm]
Random removal with $q= \frac{6}{48}$ & 48  & \textbf{3.4}/\textbf{4.5} \\[-3.0mm] \\ \hline
\end{tabular}
\end{table}

\vspace{-1mm}
\section{Experimental Results}
\label{sec:exp}

\vspace{-1mm}

\subsection{Setup}
\vspace{-1mm}
We experiment on datasets of a variety of scenarios. Table~\ref{tab:datasets} describes these datasets. We use Kaldi \cite{povey2011kaldi} for data preparation and feature extraction -- 83-dim log-mel filterbank frames with pitch \cite{ghahremani2014pitch}. SpecAugument \cite{park2019specaugment} is used on all datasets but WSJ. On all datasets, following the previous works \cite{Dong2018Speech-transformer, karita2019comparative,lu2020exploring}, we employ the \textbf{12layer Encoder-6layer Decoder} architecture. The multi-head attention components of the self-attention layers have 4 attention heads and $d^{\V}=d^{\K}=64$, $d^{\M}=256$. For the feed-forward module of the self-attention layers, $d^{\FF}=2048$. Below the encoder of the Transformer, there are two convolutional neural network layers with 256 channels, with a stride of 2 and a kernel size of 3, which map the dimension of the input sequence to $d^{\M}$. The structure of the self-attention component of the Conformer is the same as the Transformer. The kernel size of the convolution component is either 15 or 31 (depends on the dataset). Input sequences to the encoder and the decoder are concatenated with sinusoidal positional encoding \cite{vaswani2017attention,dai2019transformer}. Models are implemented using ESPnet  \cite{watanabe2018espnet} and PyTorch \cite{paszke2017automatic}.


\begin{table}[t]
\caption{Datasets Description}
\vspace{-2.5mm}
\label{tab:datasets}
\centering
\begin{tabular}{l|ccc}
\hline
datasets    & language & hours & speech style           \\ \hline
WSJ         & EN       & 81    & read                   \\
AISHELL     & ZH       & 170   & read                   \\
SWBD        & EN       & 260   &    telephone conversation   \\
AMI         & EN       & 100   & meeting                \\ \hline
\end{tabular}
\end{table}

The training schedule follows \cite{Dong2018Speech-transformer}. Dropout rate $0.1$ is used. The training lasts 100 epochs for WSJ and 50 epochs for other datasets. The averaged parameters of the last 10 epochs are used as the parameters of the final model \cite{Dong2018Speech-transformer}. Adam \cite{kingma2015adam} is used as the optimizer. The output units for WSJ/AISHELL/AMI are characters and the output tokens for SWBD experiments are tokenized at word-level using Byte Pair Encoding (BPE)~\cite{sennrich2016BPE}. The batch size is 32. Label smoothing with smoothing weight 0.1 is used. Besides the loss from the Transformer's decoder $L^{\mathrm{D}}$, a connectionist temporal classification (CTC) \cite{graves2006connectionist} loss $L^{\mathrm{CTC}}$ is also applied to the Transformer/Conformer encoder \cite{kim2017joint}. Following \cite{karita2019comparative}, the final loss $L$ for the model is: 
\vspace{-1mm}

\begin{equation}
    L = (1-\lambda)L^{\mathrm{D}} + \lambda L^{\mathrm{CTC}}.
\end{equation}
where $\lambda = 0.2$ for SWBD and $\lambda = 0.3$ for other datasets. Following our preliminary study (Table~\ref{tab:stat}), considering both the encoder and the decoder has $48$ attention heads, we test removal probability values starting from $\frac{4}{48}$.  

\vspace{-0.5mm}
\subsection{Results and discussion}
\vspace{-0.5mm}

We use WSJ to derive the proposed method and the experimental results are shown in Table~\ref{tab:stat}. Table~\ref{tab:swbd_transformer} shows the results of applying the proposed method to train Transformers on SWBD. We run the experiments with three different random seeds, and have obtained consistently reduced word error rates (WERs) with all the seeds. To further show our method has statistically significant gains, we apply Matched Pairs Sentence-Segment Word Error significance tests (mapsswe) \cite{gillick1989some} for the best Transformer models. On both SWBD/Callhome test sets, $p<0.001$.


\begin{table}[t]
\caption{WERs on SWBD for Transformer models.}
\vspace{-2.75mm}
\label{tab:swbd_transformer}
\centering
\begin{tabular}{l|c|c|c}
\hline
      & \multicolumn{3}{l}{SWBD/Callhome (WER\%)}              \\ \hline 
Removal Probability     & seed 1            & seed 2           & seed 3           \\ \hline 
\multicolumn{4}{l}{Transformer Models} \\ \hline
0/48 (baseline)    & 9.0/18.1          & 9.1/18.2          & 9.1/17.9          \\
4/48  & 8.9/17.5          & 8.7/17.6          &     -     \\
6/48  & \textbf{8.6/17.2} & \textbf{8.7/16.9}          & \textbf{8.7/16.8} \\
8/48  & 8.8/17.6 & 8.9/17.2 &  - \\
\hline
\multicolumn{3}{l|}{ Transformer \cite{karita2019comparative}} & \multicolumn{1}{c}{9.0/18.1}                  \\ \hline
\multicolumn{3}{l|}{ Very deep self-attention \cite{pham2019veryDeep}} & \multicolumn{1}{c}{10.4/18.6}                 \\ \hline
\multicolumn{3}{l|}{Multi-stride self-attention \cite{han2019multi}} & \multicolumn{1}{c}{9.1/-}                     \\ \hline
\end{tabular}
\vspace{1mm}
\end{table}


We further use our method to train Conformer models on SWBD. Noticing that in the Transformer experiments the probability values which lead to the best models are within the range of 0.1 to 0.2, we test use 0.1 or 0.2 as the removal probability. That is, we test not considering the total number of attention heads when setting the removal probability. In addition, we augment SWBD by speed perturbation \cite{ko2015audio} with ratio 0.9 and 1.1 and decode with a recurrent neural network language model (RNNLM) \cite{hori2017advances}. Table~\ref{tab:swbd_conformer} shows our method is effective in this set of experiments. We train Conformer models using a fixed random seed. The only random factor is how the attention heads are removed in each training update and the performance of the baseline is constant. We apply a two-tailed t-test to examine if the mean WERs of the five runs of the conformer models trained with removal probability 0.2 are significantly different from the baseline Conformer. On SWBD $p<0.003$ and on Callhome $p<0.02$, which shows the gains are statistically significant. Further, to the best of our knowledge, we have achieved the SOTA for end-to-end Transformer based models on SWBD.  


\begin{table}[t]
\caption{WERs on SWBD for Conformer models. * indicates the number is the average of 5 runs. In these five runs, the WERs on SWBD/Callhome ranges from 6.7 to 6.8/13.5 to 13.9.}
\vspace{-2.75mm}
\label{tab:swbd_conformer}
\centering
\begin{tabular}{l|c}
\hline
Removal Probability       & SWBD/Callhome (WER\%)             \\ \hline 
\multicolumn{2}{l}{Conformer Models} \\ \hline
0.0 (baseline) & 8.4/17.1          \\
0.1 & 8.2/\textbf{16.4}          \\
0.2 & \textbf{8.1}/\textbf{16.4}          \\ \hline
\multicolumn{2}{l}{+Speed Perturbation + RNNLM} \\ \hline
0.0 (baseline) & 6.9/14.0          \\
0.1 & \textbf{6.4} /14.0         \\
0.2 & 6.7*/\textbf{13.7}*          \\ \hline
Conformer \cite{guo2020recent} & 7.1/15.0          \\ \hline
\end{tabular}
\end{table}


Table~\ref{tab:aishell} shows the results on AISHELL. We have applied mapsswe. On the Dev set, $p<0.001$ for both the Transformer and Conformer models. On the Test set, $p=0.030$ for the Transformer and $p =0.023$ for the Conformer. We notice there is a performance gap between our Conformer models and the previous work \cite{guo2020recent}. This is because of different training configurations. We did not further tune our Conformer models on AISHELL due to the large number of experiments and we have indeed obtained statistically significant gains.


\begin{table}[t]
\caption{CERs on AISHELL of stochastic head removal.}
\vspace{-2.75mm}
\label{tab:aishell}
\centering
\begin{tabular}{l|c}
\hline
Removal Probability      & Dev/Test (CER\%)              \\ \hline 
\multicolumn{2}{l}{Transformer Models} \\ \hline
0.0 + speed perturbation + RNNLM   & 6.0/6.5                  \\ \hline
0.1 + speed perturbation + RNNLM & \textbf{5.8/6.3}                \\ \hline
0.2 + speed perturbation + RNNLM & \textbf{5.8/6.3}                \\ \hline
Transformer \cite{karita2019comparative}  & 6.0/6.7                  \\ \hline
+ 5,000 hours pretrain \cite{jiang2020further}  & -/6.26                  \\ \hline
Memory equipped self-attention \cite{gao2020san}  & 5.74/6.46                 \\  \hline \hline
\multicolumn{2}{l}{Conformer Models} \\ \hline
0.0 + speed perturbation + RNNLM   & 5.3/6.0                  \\ \hline
0.1 + speed perturbation + RNNLM   & \textbf{5.2/5.8}                  \\ \hline
Conformer \cite{guo2020recent}   & 4.4/4.7     \\ \hline
\end{tabular}
\vspace{1mm}
\end{table}


We use AMI to test our method on a more difficult ASR task. We use the individual headset microphone (IHM) setup. Table~\ref{tab:ami} shows our method has outperformed the baselines. Our method has achieved the SOTA for Transformer based models on AMI. We also applied mapsswe and $p<0.001$ for both Transformer and Conformer models on both dev and test sets. Thus, we conclude on all the datasets, our method have offered statistically significant performance gains.

We noticed that the performance of our method varies among different removal probabilities. However, the probabilities between 0.1 and 0.2 have given the best accuracies. We propose to use a development set to choose the probability or using an ensemble of models trained with different probabilities as the final model, and we left these as further works. 

\begin{table}
\caption{WERs on AMI for stochastic head removal.}
\vspace{-2.75mm}
\label{tab:ami}
\centering
\begin{tabular}{l|c}
\hline
Removal Probability & dev/test (WER \%) \\ \hline
\multicolumn{2}{l}{Transformer Models} \\ \hline 
0.0 (baseline)         &  24.9/25.8                \\ \hline
0.1                    &   \textbf{24.2/24.6}      \\ \hline
0.2                    &   24.5/24.9               \\ \hline \hline
\multicolumn{2}{l}{Conformer Models}               \\ \hline 
0.0 (baseline)         &  24.9/25.8                \\ \hline
0.1                    &   \textbf{24.1/24.2}      \\ \hline \hline
2D Conv \cite{Oglic2020}            &  24.9/24.6   \\ \hline
TDNN \cite{Oglic2020}               &  25.3/26.0   \\ \hline
\end{tabular}
\vspace{-2mm}
\end{table}

\begin{figure}[h]
\centering
\includegraphics[width=\columnwidth, height=51mm]{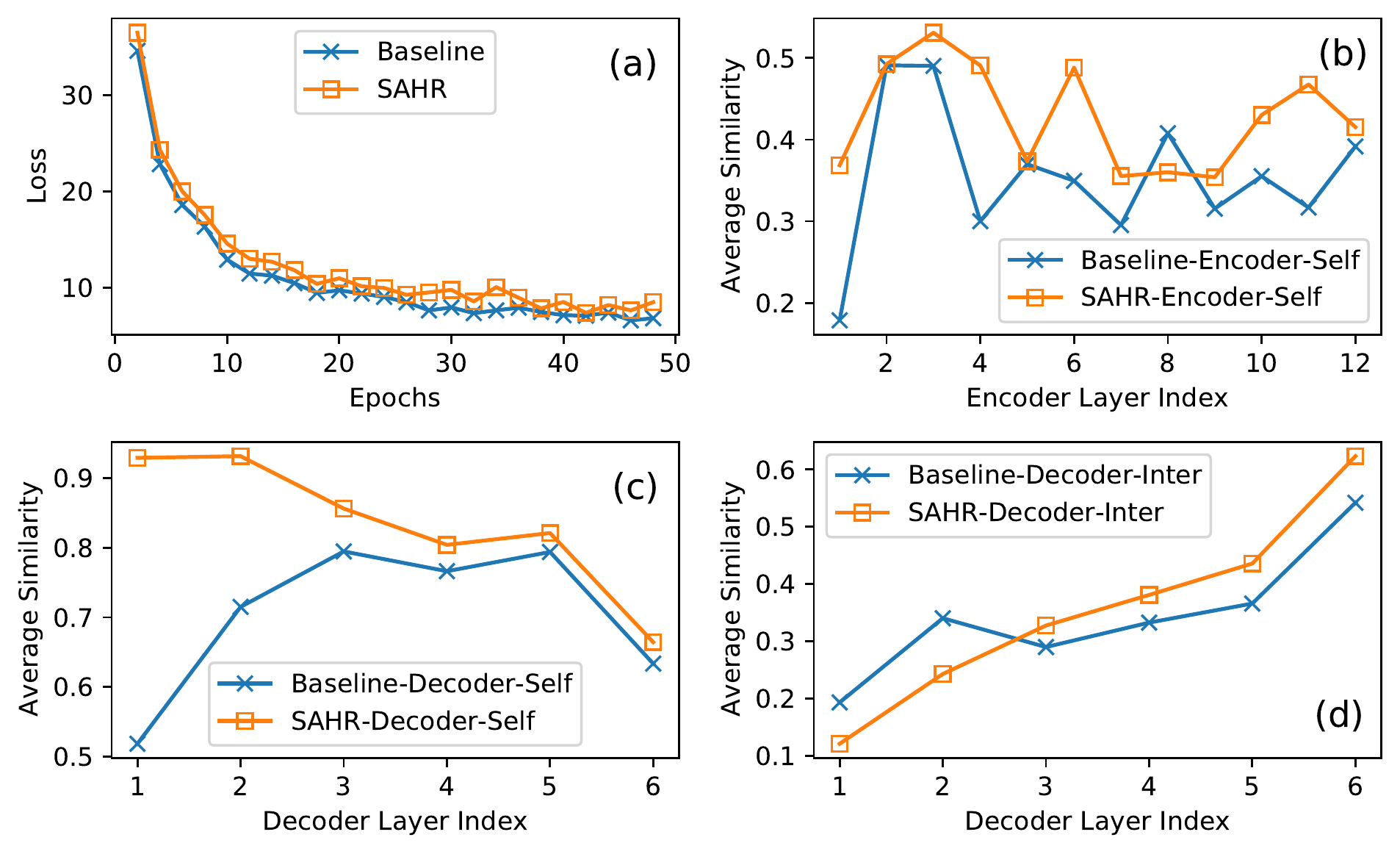}
\vspace{-5.5mm}
\caption{The train loss and the averaged similarity between attention heads of the baseline Conformer model and the Conformer model trained with stochastic attention head removal (SAHR) probability 0.1. "decoder-inter" denotes the MHA component that interacts between the encoder and the decoder. } 
\label{fig:train effect}
\end{figure}

We further study the effect of the proposed method. We have plotted the training loss of the baseline Conformer and the Conformer trained with removal probability 0.1 on SWBD. Figure~\ref{fig:train effect} (a) shows the train loss curves. The proposed method has higher train loss in each epoch. This is as excepted since in each training update only a fraction of the network is trained. We also compute the averaged similarity of the attention matrices of each attention head pair in each layer. We use the averaged cosine similarity of rows of the same indices as the similarity of attention matrices since each row vector shows how the attention weights are distributed to encode the input at the row index. Figure~\ref{fig:train effect} (b) show that with the random removal, the similarities between attention heads increase. Thus, it indicates the proposed method forces each attention head to extract the most essential patterns while allowing different heads to encode individual additional information. The baseline model has a smaller similarity between attention heads, further indicating each layer relies on some attention heads to learn the most useful features while other heads encode unessential information. Therefore, the proposed method better utilizes the representation powers of the multi-head attention and gives improved accuracies.

\vspace{-3mm}

\section{Conclusion}
\vspace{-2mm}
We observe there are redundant attention heads in Transformer based ASR models. Instead of searching for the best structure (which is time prohibitive), we propose to randomly remove attention heads during training and keeping all the attention heads at testing. The proposed method also forces each attention head to learn the key patterns, and thus better exploits the representation power of the multi-head structure. Our method has offered statistically significant improvements for Transformer and Conformer models on datasets of different ASR scenarios. Further work includes ensembling the models trained with different removal probabilities. Source code: \url{https://github.com/s1603602/attention_head_removal}

\newpage

\bibliographystyle{IEEEtran}

\bibliography{mybib}

\begin{thebibliography}{10}
\providecommand{\url}[1]{#1}
\csname url@samestyle\endcsname
\providecommand{\newblock}{\relax}
\providecommand{\bibinfo}[2]{#2}
\providecommand{\BIBentrySTDinterwordspacing}{\spaceskip=0pt\relax}
\providecommand{\BIBentryALTinterwordstretchfactor}{4}
\providecommand{\BIBentryALTinterwordspacing}{\spaceskip=\fontdimen2\font plus
\BIBentryALTinterwordstretchfactor\fontdimen3\font minus
  \fontdimen4\font\relax}
\providecommand{\BIBforeignlanguage}[2]{{%
\expandafter\ifx\csname l@#1\endcsname\relax
\typeout{** WARNING: IEEEtran.bst: No hyphenation pattern has been}%
\typeout{** loaded for the language `#1'. Using the pattern for}%
\typeout{** the default language instead.}%
\else
\language=\csname l@#1\endcsname
\fi
#2}}
\providecommand{\BIBdecl}{\relax}
\BIBdecl

\bibitem{vaswani2017attention}
A.~Vaswani, N.~Shazeer, N.~Parmar, J.~Uszkoreit, L.~Jones, A.~N. Gomez,
  {\L}.~Kaiser, and I.~Polosukhin, ``Attention is all you need,'' in
  \emph{NeurIPS}, 2017.

\bibitem{Dong2018Speech-transformer}
L.~Dong, S.~Xu, and B.~Xu, ``Speech-transformer: a no-recurrence
  sequence-to-sequence model for speech recognition,'' in \emph{ICASSP}, 2018.

\bibitem{pham2019veryDeep}
N.-Q. Pham, T.-S. Nguyen, J.~Niehues, M.~M{\"u}ller, and A.~Waibel, ``Very deep
  self-attention networks for end-to-end speech recognition,''
  \emph{INTERSPEECH}, 2019.

\bibitem{karita2019comparative}
S.~Karita, N.~Chen, T.~Hayashi, T.~Hori, H.~Inaguma, Z.~Jiang, M.~Someki,
  N.~E.~Y. Soplin, R.~Yamamoto, X.~Wang \emph{et~al.}, ``A comparative study on
  transformer vs rnn in speech applications,'' in \emph{ASRU}, 2019.

\bibitem{han2019multi}
K.~J. Han, J.~Huang, Y.~Tang, X.~He, and B.~Zhou, ``Multi-stride self-attention
  for speech recognition.'' in \emph{INTERSPEECH}, 2019.

\bibitem{jiang2020further}
D.~Jiang, W.~Li, R.~Zhang, M.~Cao, N.~Luo, Y.~Han, W.~Zou, and X.~Li, ``A
  further study of unsupervised pre-training for transformer based speech
  recognition,'' \emph{arXiv preprint arXiv:2005.09862}, 2020.

\bibitem{tian2020synchronous}
Z.~Tian, J.~Yi, Y.~Bai, J.~Tao, S.~Zhang, and Z.~Wen, ``Synchronous
  transformers for end-to-end speech recognition,'' in \emph{ICASSP}, 2020.

\bibitem{bai2020laso}
Y.~Bai, J.~Yi, J.~Tao, Z.~Tian, Z.~Wen, and S.~Zhang, ``Listen attentively, and
  spell once: Whole sentence generation via a non-autoregressive architecture
  for low-latency speech recognition,'' in \emph{INTERSPEECH}, 2020.

\bibitem{luo2020simplified}
H.~Luo, S.~Zhang, M.~Lei, and L.~Xie, ``Simplified self-attention for
  transformer-based end-to-end speech recognition,'' \emph{arXiv preprint
  arXiv:2005.10463}, 2020.

\bibitem{gulati2020conformer}
A.~Gulati, J.~Qin, C.-C. Chiu, N.~Parmar, Y.~Zhang, J.~Yu, W.~Han, S.~Wang,
  Z.~Zhang, Y.~Wu \emph{et~al.}, ``Conformer: Convolution-augmented transformer
  for speech recognition,'' in \emph{INTERSPEECH}, 2020.

\bibitem{zhang2021usefulness}
S.~Zhang, E.~Loweimi, P.~Bell, and S.~Renals, ``On the usefulness of
  self-attention for automatic speech recognition with transformers,'' in
  \emph{IEEE SLT}, 2021.

\bibitem{voita2019analyzing}
E.~Voita, D.~Talbot, F.~Moiseev, R.~Sennrich, and I.~Titov, ``Analyzing
  multi-head self-attention: Specialized heads do the heavy lifting, the rest
  can be pruned,'' in \emph{ACL}, 2019.

\bibitem{paul1992wsj}
D.~B. Paul and J.~M. Baker, ``The design for the wall street journal-based csr
  corpus,'' in \emph{Proceedings of the workshop on Speech and Natural
  Language}.\hskip 1em plus 0.5em minus 0.4em\relax Association for
  Computational Linguistics, 1992.

\bibitem{bu2017aishell}
H.~Bu, J.~Du, X.~Na, B.~Wu, and H.~Zheng, ``Aishell-1: An open-source mandarin
  speech corpus and a speech recognition baseline,'' in \emph{O-COCOSDA}, 2017.

\bibitem{godfrey1992switchboard}
J.~{Godfrey}, E.~{Holliman}, and J.~{McDaniel}, ``Switchboard: telephone speech
  corpus for research and development,'' in \emph{ICASSP}, 1992.

\bibitem{renals2007recognition}
S.~Renals, T.~Hain, and H.~Bourlard, ``Recognition and understanding of
  meetings the ami and amida projects,'' in \emph{ASRU}, 2007.

\bibitem{srivastava2014dropout}
N.~{Srivastava}, G.~{Hinton}, A.~{Krizhevsky}, I.~{Sutskever}, and
  R.~{Salakhutdinov}, ``Dropout: a simple way to prevent neural networks from
  overfitting,'' \emph{JMLR}, 2014.

\bibitem{michel2019are}
P.~{Michel}, O.~{Levy}, and G.~{Neubig}, ``Are sixteen heads really better than
  one,'' in \emph{NeurIPS}, 2019.

\bibitem{fan2019reducing}
A.~Fan, E.~Grave, and A.~Joulin, ``Reducing transformer depth on demand with
  structured dropout,'' in \emph{ICLR}, 2019.

\bibitem{peng2020mixture}
H.~Peng, R.~Schwartz, D.~Li, and N.~A. Smith, ``A mixture of $ h-1$ heads is
  better than $ h $ heads,'' \emph{ACL}, 2020.

\bibitem{zhou2020scheduled}
W.~Zhou, T.~Ge, K.~Xu, F.~Wei, and M.~Zhou, ``Scheduled drophead: A
  regularization method for transformer models,'' in \emph{EMNLP}, 2020.

\bibitem{wu2020lite}
Z.~Wu, Z.~Liu, J.~Lin, Y.~Lin, and S.~Han, ``Lite transformer with long-short
  range attention,'' in \emph{ICLR}, 2020.

\bibitem{ba2016layer}
J.~L. Ba, J.~R. Kiros, and G.~E. Hinton, ``Layer normalization,'' \emph{arXiv
  preprint arXiv:1607.06450}, 2016.

\bibitem{hochreiter1997long}
S.~Hochreiter and J.~Schmidhuber, ``Long short-term memory,'' \emph{Neural
  computation}, 1997.

\bibitem{povey2011kaldi}
D.~Povey, A.~Ghoshal, G.~Boulianne, L.~Burget, O.~Glembek, N.~Goel,
  M.~Hannemann, P.~Motlicek, Y.~Qian, P.~Schwarz \emph{et~al.}, ``The {Kaldi}
  speech recognition toolkit,'' in \emph{ASRU}, 2011.

\bibitem{ghahremani2014pitch}
P.~{Ghahremani}, B.~{BabaAli}, D.~{Povey}, K.~{Riedhammer}, J.~{Trmal}, and
  S.~{Khudanpur}, ``A pitch extraction algorithm tuned for automatic speech
  recognition,'' in \emph{ICASSP}, 2014.

\bibitem{park2019specaugment}
D.~S. Park, W.~Chan, Y.~Zhang, C.-C. Chiu, B.~Zoph, E.~D. Cubuk, and Q.~V. Le,
  ``Specaugment: A simple data augmentation method for automatic speech
  recognition,'' \emph{INTERSPEECH}, 2019.

\bibitem{lu2020exploring}
L.~Lu, C.~Liu, J.~Li, and Y.~Gong, ``Exploring transformers for large-scale
  speech recognition,'' in \emph{INTERSPEECH}, 2020.

\bibitem{dai2019transformer}
Z.~{Dai}, Z.~{Yang}, Y.~{Yang}, J.~G. {Carbonell}, Q.~V. {Le}, and
  R.~{Salakhutdinov}, ``Transformer-xl: Attentive language models beyond a
  fixed-length context.'' in \emph{ACL}, 2019.

\bibitem{watanabe2018espnet}
S.~Watanabe, T.~Hori, S.~Karita, T.~Hayashi, J.~Nishitoba, Y.~Unno, N.~E.~Y.
  Soplin, J.~Heymann, M.~Wiesner, N.~Chen, A.~Renduchintala, and T.~Ochiai,
  ``{ESPnet}: End-to-end speech processing toolkit,'' \emph{INTERSPEECH}, 2018.

\bibitem{paszke2017automatic}
A.~Paszke, S.~Gross, S.~Chintala, G.~Chanan, E.~Yang, Z.~DeVito, Z.~Lin,
  A.~Desmaison, L.~Antiga, and A.~Lerer, ``Automatic differentiation in
  pytorch,'' in \emph{NeurIPS 2017 Workshop Autodiff}, 2017.

\bibitem{kingma2015adam}
D.~P. {Kingma} and J.~L. {Ba}, ``Adam: A method for stochastic optimization,''
  in \emph{ICLR}, 2015.

\bibitem{sennrich2016BPE}
R.~{Sennrich}, B.~{Haddow}, and A.~{Birch}, ``Neural machine translation of
  rare words with subword units,'' in \emph{ACL}, 2016.

\bibitem{graves2006connectionist}
A.~{Graves}, S.~{Fernández}, F.~{Gomez}, and J.~{Schmidhuber}, ``Connectionist
  temporal classification: labelling unsegmented sequence data with recurrent
  neural networks,'' in \emph{ICML}, 2006.

\bibitem{kim2017joint}
S.~Kim, T.~Hori, and S.~Watanabe, ``Joint ctc-attention based end-to-end speech
  recognition using multi-task learning,'' in \emph{ICASSP}, 2017.

\bibitem{gillick1989some}
L.~{Gillick} and S.~{Cox}, ``Some statistical issues in the comparison of
  speech recognition algorithms,'' in \emph{ICASSP}, 1989.

\bibitem{ko2015audio}
T.~Ko, V.~Peddinti, D.~Povey, and S.~Khudanpur, ``Audio augmentation for speech
  recognition,'' in \emph{INTERSPEECH}, 2015.

\bibitem{hori2017advances}
T.~Hori, S.~Watanabe, Y.~Zhang, and W.~Chan, ``Advances in joint ctc-attention
  based end-to-end speech recognition with a deep cnn encoder and rnn-lm,'' in
  \emph{INTERSPEECH}, 2017.

\bibitem{guo2020recent}
P.~{Guo}, F.~{Boyer}, X.~{Chang}, T.~{Hayashi}, Y.~{Higuchi}, H.~{Inaguma},
  N.~{Kamo}, C.~{Li}, D.~{Garcia-Romero}, J.~{Shi}, J.~{Shi}, S.~{Watanabe},
  K.~{Wei}, W.~{Zhang}, and Y.~{Zhang}, ``Recent developments on espnet toolkit
  boosted by conformer.'' \emph{arXiv preprint arXiv:2010.13956}, 2020.

\bibitem{gao2020san}
Z.~Gao, S.~Zhang, M.~Lei, and I.~McLoughlin, ``San-m: Memory equipped
  self-attention for end-to-end speech recognition,'' in \emph{INTERSPEECH},
  2020.

\bibitem{Oglic2020}
D.~Oglic, Z.~Cvetkovic, P.~Bell, and S.~Renals, ``A deep 2d convolutional
  network for waveform-based speech recognition,'' in \emph{INTERSPEECH}, 2020.

\end{thebibliography}


\end{document}